%% file: root.tex
\documentclass[letterpaper, 10 pt, conference]{ieeeconf}  

\IEEEoverridecommandlockouts                              

\usepackage{times}
\usepackage{amsmath,amsfonts}
\usepackage{mathtools}
\usepackage{float}
\usepackage{textcomp}
\usepackage{multicol}
\usepackage{multirow}
\usepackage{balance}
\usepackage[short,c2]{optidef}

\usepackage{makeidx}  
\makeindex
\usepackage{xcolor}
\usepackage{colortbl}
\usepackage{placeins}
\usepackage{booktabs}
\usepackage{siunitx}
\usepackage{algorithm}
\usepackage{algpseudocode}
\usepackage{bm} 
\usepackage{cuted} 
\usepackage{capt-of}
\usepackage{subcaption} 
\usepackage{xcolor}
\usepackage[normalem]{ulem}
\newenvironment{remark}
  {\par\noindent\textbf{Remark.}\ }
  {\par}

\definecolor{mycitecolor}{RGB}{71, 191, 38}
\definecolor{mylinkcolor}{RGB}{40, 115, 201}
\makeatletter
\let\NAT@parse\undefined
\makeatother
\usepackage[bookmarks=true, colorlinks, citecolor=mycitecolor,linkcolor=mylinkcolor,urlcolor=mycitecolor]{hyperref}

\newcommand{\maulik}[1]{\textcolor{black}{#1}}

\title{\LARGE 
\textbf{MIMIC-D}: \textbf{M}ulti-modal \textbf{I}mitation for \textbf{M}ult\textbf{I}-agent \textbf{C}oordination with \textbf{D}ecentralized Diffusion Policies
\vspace{-0.4cm}
}

\author{Dayi Dong\textsuperscript{*}, Maulik Bhatt\textsuperscript{*}, Seoyeon Choi, and Negar Mehr
\thanks{All authors are with the Department of Mechanical Engineering, University of California Berkeley, Berkeley, CA 94709, USA {\tt \small \{dayi.dong, maulikbhatt, seoyeon99, negar\}@berkeley.edu}}%
\thanks{This work was supported by the National Science Foundation under Grants ECCS-2438314 (CAREER Award), CNS-2529645, and CCF-2423134, and by the Army Research Laboratory under Grant W911NF-26-1-0002.
}
}

\begin{document}

\maketitle

\begingroup
\renewcommand\thefootnote{\fnsymbol{footnote}}
\footnotetext[1]{Indicates equal contribution.}
\endgroup

\thispagestyle{empty}
\pagestyle{empty}

\begin{strip}
  \centering
  \vspace{-2.6cm}
  \captionsetup{type=figure,hypcap=false}
  \includegraphics[width=\textwidth]{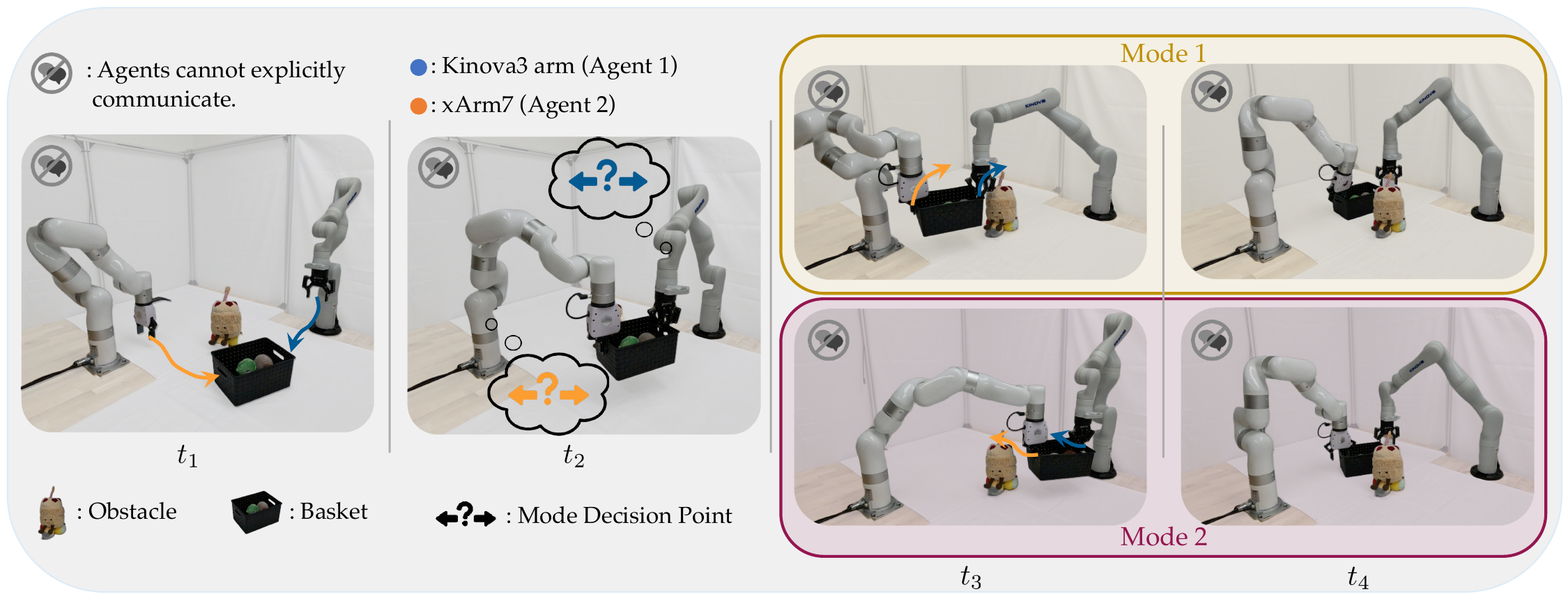}
  \captionof{figure}{\textbf{MIMIC-D deployed on a bimanual manipulation setup.} An xArm7 and a Kinova3 robotic arm collaborate to lift a basket around an obstacle. The task presents a multi-modal coordination challenge as the arms need to coordinate to either pass the obstacle on the right or on the left. Using our method, the arms achieve coordination by independently sampling their policies based on local observations without explicit communication. Our method successfully recovers both solution modes, demonstrating its ability to capture diverse coordination strategies while avoiding the freezing robot problem. The panels depict the starting configuration ($t_1$), mode decision ($t_2$), and the completion of both modes ($t_3\, , t_4$).}
  \label{fig:fig1}
  \vspace{-0.4cm}
\end{strip}


\begin{abstract}

As robots become more integrated in society, their ability to coordinate with other robots and humans on multi-modal tasks (those with multiple valid solutions) is crucial. Such behaviors can be learned from expert demonstrations via imitation learning (IL), but when expert demonstrations are multi-modal, standard IL approaches usually average across modes or collapse to a single mode, preventing effective coordination. 
Being inspired by diffusion models’ ability to capture complex multi-modal trajectory distributions in single-agent settings, we develop a diffusion-based framework for coordinated multi-modal behavior in multi-agent systems.
However, existing multi-agent diffusion approaches typically require a centralized planner or explicit communication among agents. This assumption can fail in real-world scenarios where robots must operate independently or with agents like humans that they cannot directly communicate with. Therefore, we propose MIMIC-D, a \maulik{joint} training with decentralized execution paradigm for multi-modal multi-agent IL via diffusion. We jointly train all agents' policies with \maulik{only local information} to achieve implicit coordination. In simulation and hardware experiments, our method exhibits robust multi-modal coordination behavior in various tasks and environments, improving upon state-of-the-art baselines.

\end{abstract}


\input{introduction}

\input{related_work}

\input{ctde}

\input{experiments}

\input{conclusion}
\vspace{-0.2cm}
\FloatBarrier
\bibliographystyle{IEEEtran}
\bibliography{references}

\end{document}

%% file: introduction.tex
\vspace{-0.1cm}
\section{Introduction}
\vspace{-0.2cm}
\label{sec:intro}

    As society actively embraces new robotic developments in our everyday lives, these robots must learn to coordinate with humans and other agents in shared spaces where multiple coordination strategies may exist. For example, two people approaching head-on can avoid collision by both yielding right or both yielding left. Either strategy is acceptable, but the agents need to decide together to achieve the desired \emph{coordination} because contradicting strategies lead to failure. In many scenarios, there is no reliable central planner available, so robots must implicitly coordinate multi-modal behaviors. In this paper, we propose to learn multi-modal coordination policies from expert data of multi-agent interactions, allowing robots to perform robust, decentralized execution without explicit communication.
   
    Many approaches for solving this multi-agent coordination problem have been proposed. Learning-based methods have propelled this field forward by introducing reinforcement learning (RL)~\cite{sutton1998reinforcement} and temporal-difference (TD) methods~\cite{sutton1988learning} when it is feasible to write an explicit reward function. Imitation learning (IL) methods like behavior cloning (BC)~\cite{pomerleau1988alvinn} and generative adversarial imitation learning (GAIL)~\cite{ho2016generative} instead utilize expert demonstrations to learn a desired behavior. These methods, however, suffer from poor performance in complex long-horizon tasks~\cite{ross2011reduction, bhattacharyya2018multi, lee2021adversarial}, and struggle with multi-modal expert data, where multiple valid solution modes exist~\cite{peters2020inference,kavuncu2021potential,mehr2023maximum,bhatt2025strategic} due to mode-averaging (learning to output the average of distinct strategies) and mode collapse (perfectly imitating only one of the expert's behaviors). 
    
   Diffusion models~\cite{ho2020denoising} are a more recent approach in IL, originally developed for image generation because of their ability to model complex multi-modal data distributions. More recently, diffusion policy models~\cite{janner2022planning,chi2023diffusion} have been developed to model expert behaviors for robotics. This approach is particularly compelling because it can capture multi-modality present in expert demonstrations, which is essential for successfully manipulating objects~\cite{janner2022planning,chi2023diffusion}.
    
     We argue that \emph{multi-modality is even more common in multi-agent systems} and more essential because if you cannot capture all of the modes, there is a greater chance of task failure. Therefore, leveraging the multi-modal capabilities of diffusion models for multi-agent systems is even more essential to ensure robust multi-agent coordination. While diffusion models have been studied in the context of multi-agent systems~\cite{chi2023diffusion,jiang2023motiondiffuser}, most approaches assume a centralized framework where agents' observations and actions are jointly available during the task execution~\cite{jiang2023motiondiffuser}. However, many real-world deployments, like driving or human-robot interactions, are inherently decentralized because they lack a concrete way to communicate information, making a centralized policy unrealistic.

    To address this, we propose \textbf{MIMIC-D}: \textbf{M}ulti-modal \textbf{I}mitation for \textbf{M}ult\textbf{I}-agent \textbf{C}oordination with \textbf{D}ecentralized Diffusion Policies (see Fig.~\ref{fig:architecture}), a \maulik{joint} training with decentralized execution framework for multi-modal multi-agent diffusion-based planning. Our approach ensures that agents can implicitly coordinate without explicit communication. We propose training all the agents' policies \maulik{jointly} so that \maulik{they can} build up an understanding of how to respond to one another. Then, during execution, each agent only needs its own policy and its local observations to plan out its actions.
    Through extensive investigations in navigation and bimanual manipulation tasks, like in Fig~\ref{fig:fig1}, we find that MIMIC-D achieves significantly lower collision rates and higher task completion rates due to its ability to foster coordination among agents.
    Additionally, we can better capture the distribution of multi-modal expert trajectories than baseline methods. In summary, our contributions are as follows: 
    \begin{enumerate}
        \item A \maulik{joint training with decentralized execution} framework for multi-agent coordination using only local observations that effectively captures multi-modality in the expert data.
        \item On two and three-agent 2D navigation and simulated bimanual manipulation tasks, compared to other approaches, we achieve fewer collisions and better replication of the expert demonstration distribution.
        \item On a hardware bimanual manipulation task that requires coordination to move a basket, MIMIC-D achieves 95\% success in 20 trials.
    \end{enumerate}

%% file: related_work.tex
\begin{figure*}[htbp]
    \centering
    \includegraphics[width=\textwidth]{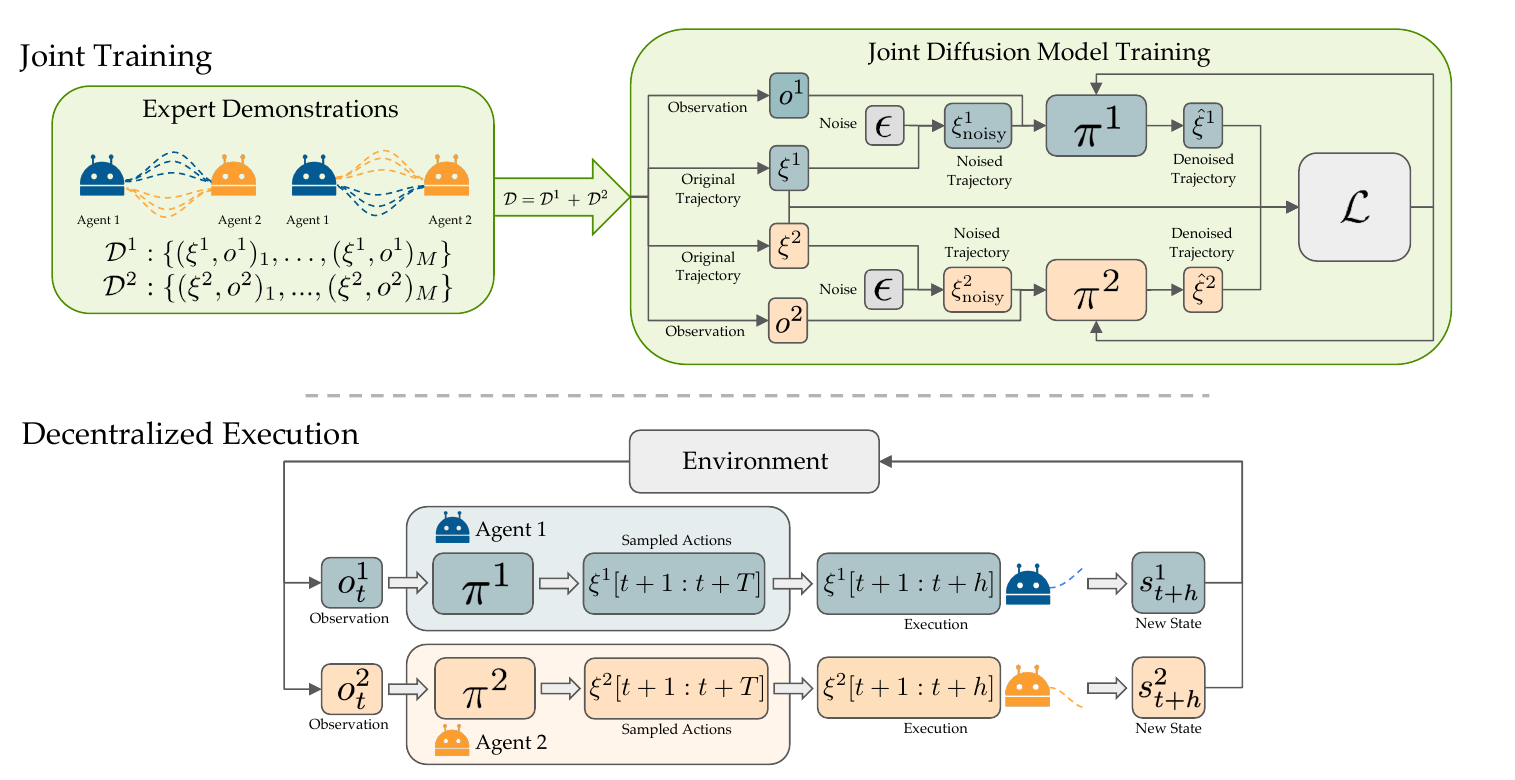}
    \caption{\textbf{An overview of our MIMIC-D framework.} In the training process (top), we utilize a dataset of multi-agent expert demonstrations to train the robot policies jointly. During the decentralized execution process (bottom), each agent plans their trajectory independently by only making use of its local observations to sample the diffusion model.}
    \label{fig:architecture}
    \vspace{-0.5cm}
\end{figure*}

\vspace{-0.2cm}
\section{Related Work} \label{sec:related}

    \subsubsection{Multi-agent Reinforcement Learning (MARL)}

    In multi-agent reinforcement learning (MARL), agents learn policies through interaction. Methods like Independent Q-Learning \cite{tan1993multi} treat other agents as part of the environment, which can lead to non-stationary and instability~ \cite{hernandez2019survey}. Centralized Training with Decentralized (CTDE) methods (e.g., MADDPG \cite{lowe2017multi},  QMIX \cite{rashid2020monotonic}, MAPPO \cite{yu2022surprising}) can stabilize the learning process by using centralized critics during training and decentralized actors during execution, but they still require manually constructable reward functions for all the agents and extensive reward shaping~\cite{nagpal2025leveraging,choi2025craft}.

    \subsubsection{Multi-agent Imitation Learning (MAIL)}
    Multi-agent imitation learning instead leverages expert demonstrations to guide the planning of individual agents. Approaches like behavior cloning (BC) \cite{pomerleau1988alvinn,ross2011reduction} are simple but suffer from compounding errors in long-horizon tasks and fail to model inter-agent dependencies required for coordination.
    Generative adversarial IL (GAIL) \cite{ho2016generative}, and by extension, multi-agent GAIL (MAGAIL) \cite{song2018multi}, can avoid significant compounding error but exhibit mode collapse when demonstrations are multi-modal. 

    Other than choosing an IL objective, selecting the training paradigm also impacts performance. Several works adopt the CTDE framework to address the MAIL problem (e.g., MisoDICE\cite{bui2025misodice}) by leveraging joint trajectories during training. However, these methods learn policies by matching the expert's state-action occupancy measure, resulting in nearly deterministic policies at execution time and limiting recovery of multi-modal behavior. We instead seek to adopt a generative approach to learn trajectory distributions that preserves the multi-modal nature of the expert demonstrations.   
    
    \subsubsection{Diffusion}
    Diffusion has shown great promise for policy learning~\cite{janner2022planning,chi2023diffusion}. Denoising Diffusion Probabilistic Models (DDPM) \cite{ho2020denoising} are the core framework in diffusion models, but additional works have also demonstrated smooth robot behavior and strong multi-modality (e.g., Diffusion Policy~\cite{chi2023diffusion}). These works primarily assume a single agent and access to the full state during execution.
    
    Recent works extend diffusion to the multi-agent setting by explicitly modeling interactions and performing centralized execution by sampling a joint multi-agent trajectory distribution (e.g., MotionDiffuser~\cite{jiang2023motiondiffuser}, DJINN~\cite{niedoba2023diffusion}). MADiff \cite{zhu2024madiff} and LatentToM \cite{he2025latent} are two works that adapt diffusion models towards a CTDE framework. MADiff utilizes teammate/opponent modeling and an attention-based diffusion model to handle interactions among agents, thereby achieving coordination. LatentToM introduces a consensus embedding that is based on shared observations from a third-party camera to align beliefs, acting as a bridge between the diffusion models. In contrast, we learn decentralized diffusion policies only conditioned on local observations during execution while still capturing the multi-modality inherent in interactions.

%% file: ctde.tex
\section{Problem Formulation}\label{sec:prob} 
    We study the problem of learning multi-modal policies for multi-agent coordination from expert data. Let $N$ be the number of agents and $\mathcal S = \mathcal{S}^1 \times \ldots \times \mathcal{S}^N$ be the joint state space where $\mathcal{S}^i \subseteq \mathbb{R}^{n_i}$ is the state space for agent $i$ with $n_i$ being the dimension of the state space of agent $i$. Let $\mathcal A = \mathcal{A}^1 \times \ldots \times \mathcal{A}^N$ be the joint action space, and $\mathcal{A}^i \subseteq \mathbb{R}^{m_i}$ is the action space for agent $i$ with $m_i$ being the dimension of the action-space for agent $i$ and $\mathcal O = \mathcal{O}^1 \times \ldots \times \mathcal{O}^N$ be the joint observation space where $\mathcal{O}^i \subseteq \mathbb{R}^{\mathrm{o}_i}$ is the observation space for agent $i$ with $\mathrm{o}_i$ being the dimension of the observation-space for agent $i$. We define $P: \mathcal{S} \times \mathcal{A} \rightarrow \mathcal{P}( \mathcal{S})$ to be the state transition function, which defines the probability distribution over the next joint state given the current joint state and action, where $\mathcal{P}( \mathcal{S})$ is the set of all probability measures on $\mathcal{S}$. Let $Z^i: \mathcal{S} \rightarrow \mathcal{P}( \mathcal{O}^i)$ be the observation function where $\mathcal{P}( \mathcal{O}^i)$ is the set of all probability measures on $\mathcal{O}^i$. The observation function $Z^i$ defines the probability of agent $i$ receiving an observation $o^i$ from the joint state $s = (s^1, \ldots, s^N)$.
    
    We model the multi-agent coordination problem as a decentralized Partially Observed Markov Decision Process (dec-POMDP) denoted via $\langle \mathcal S,\mathcal A,\mathcal O, P, Z, N\rangle$.
    At each time $t$, the system is in state $s_t\in\mathcal S$, and each agent $i$ receives a local observation $o_t^i \in \mathcal{O}^i$ and chooses an action $a_t^i \in \mathcal A^i$ according to a decentralized policy $\pi^i(a_t^i|o_t^i)$. The joint action $a_t=(a_t^1,\dots,a_t^N)$ of all the agents yields the next joint state $s_{t+1} \sim P(s_t,a_t)$.

    \subsection{Multi-Agent Imitation}

    We aim to achieve decentralized multi-modal multi-agent coordination through multi-agent imitation learning. We assume that we have access to a dataset $\mathcal{D}$ containing $M$ expert demonstrations where each demonstration is a collection of $N$ agents interacting with each other for a finite horizon of time $T$. Each demonstration in the dataset is a set of tuples $\{(\xi^i,o^i)\}_{i=1}^N$, where $o^i$ is the observation of agent $i$ and $\xi^i = \{a^i_0,\ldots,a^i_{T-1}\}$ is the corresponding finite-horizon ($T$) trajectory of actions executed by agent $i$ associated with observation $o^i$. Note that the expert demonstrations may be multi-modal in nature, i.e., for a given observation at time $t$, $o^i_t$, we may have two or more different expert actions. We assume that the agents exhibit coordination through various modes in the expert demonstrations of interactions.

    We parameterize each agent $i$'s policy via parameters $\theta^i$. Let $\theta = \{\theta^i\}_{i=1}^N$ be the set of joint policy parameters of all agents. Our goal is to learn a set of decentralized policies $\{\pi_{\theta^1}^1, \ldots, \pi_{\theta^N}^N\}$ that can collectively reproduce individual agents' behaviors and learn implicit coordination among the agents from the dataset $\mathcal{D}$. To achieve this, we operate under the \maulik{joint training with decentralized execution} paradigm. During the \maulik{joint} training, the agents share a common loss function, allowing our learning process to learn coordination among agents. During decentralized execution, we want each policy to operate independently, where each agent $i$ only relies on its local observations at time $t$, $o^i_t$, to select its action $a^i_t$. 
    
    We model each decentralized policy as a conditional diffusion model. The following section briefly reviews the continuous-time diffusion framework that forms the foundation of our approach.

\subsection{Conditional Diffusion Models}

In this section, we present the formulation of each agent $i$'s policy as a diffusion model. 
Given the observation $o^i$ for each agent $i$, we assume that the expert action trajectories are distributed according to an unknown distribution $\xi^i \sim p_{\text{data}}(\xi^i;o^i)$ with standard deviation $\sigma^i_{\text{data}}$. Conditional Diffusion models \cite{karras2022elucidating} are generative models that aim to generate samples from this unknown distribution, and achieve this by considering a family of modified distributions $p(\xi^i;\sigma^i,o^i)$ obtained by iteratively adding i.i.d. Gaussian noise of standard deviation $\sigma^i$ to the data. For $\sigma^i_{\text{max}} >> \sigma^i_{\text{data}}$, $p(\xi^i;\sigma^i_{\text{max}},o^i)$ is practically indistinguishable from pure Gaussian noise. In diffusion models, we learn a denoising process that parametrizes the conditional probability distribution $p_{\theta^i}(\xi^i;o^i)$ of trajectories via a learned denoiser $D_{\theta^i}(\xi^i; \sigma^i, o^i)$ parameterized by $\theta^i$. The idea is to randomly sample a noisy trajectory $\xi^i_0 \sim \mathcal{N}(0,{\sigma^{i^2}_\text{max}}\mathbf{I})$ and sequentially denoise it into trajectories $\xi^i_k$ with noise levels $\sigma^i_{0} = \sigma^i_{\text{max}} > \sigma^i_1 > \cdots > \sigma^i_K = 0$ such that $\xi^i_K$ is distributed according to the original data distribution $p_{\text{data}}$. 

The denoiser function, $D_{\theta^i}( \xi^i ; \sigma^i, o^i)$, is usually a neural network trained to predict a clean trajectory $\xi^i$ from a noisy version:
\vspace{-0.3cm}
\begin{equation}\label{eq:add-noise}
    \xi^i_{\text{noisy}} = \xi^i + \epsilon,
\end{equation} 
where $\epsilon \sim \mathcal{N}(0, {\sigma^i}^2\mathbf{I})$.
The model is trained by optimizing a simple denoising objective. We sample a noise level $\sigma^i$ for agent $i$ according to $p_\text{train}(\sigma^i)$ defined by $\ln(\sigma^i) \sim \mathcal{N}(P_{\text{mean}}, P_{\text{std}}^2)$, as done in \cite{karras2022elucidating}, and noise $\epsilon \sim \mathcal{N}(0, {\sigma^i}^2\mathbf{I})$ and minimize the expected error:
\begin{equation} \label{eq:loss}
    \mathcal L^i_{\text{diff}}(\theta^i)
    =\mathbb E_{\substack{\xi^i \sim p_{_{\text{data}}},\;\sigma^i \sim p_\text{train}(\sigma^i)\\ \epsilon \sim \mathcal N(0,{\sigma^i}^2\mathbf I)}}
    \Big[\big\| D_{\theta^i}(\,\xi^i+\epsilon;\sigma^i,o^i\,) - \xi^i \big\|_2^2 \Big].
\end{equation}

To generate a new trajectory, we solve a probability flow ODE that transforms a sample from a simple noise distribution into a sample from the data distribution. The process starts with a sample drawn from pure Gaussian noise, $\xi^i_0 \sim \mathcal{N}(0, {\sigma^i_{\text{max}}}^2\mathbf{I})$, and integrates backwards from $\sigma^i_{\text{max}}$ to 0. 
We use Algorithm 1 from \cite{karras2022elucidating} to solve this ODE numerically and obtain the denoised trajectory $\xi^i_K$.

While early diffusion models commonly used U-Net or Convolutional Neural Net backbones \cite{ho2020denoising,nichol2021improved,dhariwal2021diffusion,rombach2022high}, recent work has instead leveraged a transformer backbone (DiT) \cite{peebles2023scalable}. The implementation of a transformer results in better scaling, improved long-horizon task performance, and flexibility in conditioning for a diffusion model. We will follow the same architecture for our diffusion policies and in the following section, detail our MIMIC-D method.

\section{
{MIMIC-D} 
} \label{subsec:method}
In our approach, we model decentralized policies $\pi^i$ as a conditional diffusion model with denoiser network $D_{\theta^i}(\xi^i;\sigma^i, o^i)$.  
For each agent, $D_{\theta^i}$ takes three inputs, the noisy action trajectory $\xi^i_k$, current noise level $\sigma^i_k$, and observation $o^i$,
and with iterative denoising produces a sampled action trajectory ${\xi}^i_K$. The observation $o^i$ helps give context to the policy at planning time and encodes information about the ego agent, other agents in the system, and the relevant context it can observe. In the following section, we describe the \maulik{training} aspects of our MIMIC-D method. An overview of the MIMIC-D framework is provided in Fig.~\ref{fig:architecture}.

\subsection{\maulik{Joint} Training} \label{subsubsec:training}
We jointly train the policies for all the agents in the system. During the training, \maulik{the agents' denoising policies share a single loss and their local observations contain information about the other agents as well}. This is necessary to promote coordination and collision-avoidant behaviors by encouraging agents to learn the influence of their actions over others.

The training dataset $\mathcal{D}$ consists of $M$ joint expert demonstrations, each of horizon $T$. 
We train a set of distinct policies $\pi_\theta = \{\pi_{\theta^1}^1, \ldots, \pi_{\theta^N}^N\}$ such that each agent has an individual denoising policy. Therefore, each agent $i$ has its own unique parameters $\theta^i$ and denoising loss from \eqref{eq:loss}. In the \maulik{joint} training process, we define our total shared loss to be
\vspace{-0.3cm}
\begin{equation}\label{eq:total_loss}
    \mathcal{L}_{\text{total}}(\theta) = \sum_{i=1}^N \mathcal{L}^i_{\text{diff}}(\theta^i)
\end{equation}
\vspace{-0.4cm}

\noindent as the sum of the losses for each agent.

We present the training loop in Algorithm~\ref{alg:training}. At each gradient step, we loop over agents, sum the per-agent diffusion losses, then update $\theta$ by a single AdamW step using the gradient of our total loss with respect to all parameters $\theta$.
This architecture trains a specialized policy for each agent, which allows for heterogeneity among different agents in the same system. Additionally, since we train the policy parameters using the joint loss function, it encourages agents to coordinate. The shared optimization step adjusts parameters based on the entire system's performance, which encourages collaborative actions to achieve shared goals.

\begin{algorithm}
\caption{MIMIC-D Training Procedure}\label{alg:training}
\begin{algorithmic}[1]
\State \textbf{Input:} Expert dataset $\mathcal{D}$ 
\State \textbf{Initialize:} Policy parameters $\theta = \{\theta^1, \ldots, \theta^N\}$ and a joint optimizer \texttt{Optim}

\For{number of training steps}
    \State $\mathcal{L}_{\text{total}} \gets 0$
    \For{$i = 1, \ldots, N$} \Comment{Loop through each agent}
        \State Sample batch ${\xi}^i, {o^i} \sim \mathcal{D},$
        \State Sample noise levels $\sigma^i \sim p_{\text{train}}(\sigma^i)$
        \State Create noisy batch ${\xi^i_{\text{noisy}}}$ using \eqref{eq:add-noise}
        \State Predict denoised batch $\hat{{\xi}}^i \gets D_{\theta^i}({\xi^i_{\text{noisy}}}; {\sigma^i}, {o^i})$
        \State Compute loss $\mathcal{L}^i_{\text{diff}}(\theta^i)$ according to \eqref{eq:loss}
        \State $\mathcal{L}_{\text{total}} \gets \mathcal{L}_{\text{total}} + \mathcal{L}^i_{\text{diff}}(\theta^i)$
    \EndFor
    \State Update all parameters $\theta \leftarrow \texttt{Optim}\!\big(\theta,\,\nabla_{\theta}\mathcal{L}_{\text{total}}\big)$ 

\EndFor
\end{algorithmic}
\end{algorithm}

\subsection{Decentralized Online Execution} \label{subsubsec:execution}
Classical diffusion policy makes use of a single centralized policy that is aware of all agents' histories and future plans at execution time. Instead, our trained policies can be executed in a decentralized fashion with each agent having access to only local observations. 
Additionally, we implement our approach in a receding time horizon manner to allow for replanning that considers the current positions of the other agents in the environment. This gives our method the ability to have agents adjust to changes in the environment and to adapt to one another's actions. At every timestep, each agent $i$ individually acquires their local observation $o^i$ from the environment. Then, the agents sample their own policy $\pi^i_{\theta^i}$ for ${\xi}^i$, which has horizon $T$. Only the first $h$ ($h < T$) steps of each agent's actions are executed. This observing-sampling-executing process is then repeated. 




%% file: experiments.tex
\section{Simulated Experiments} \label{sec:experiments}
    In this section, we detail our simulation setups, baselines, and evaluation metrics used to validate MIMIC-D's ability to coordinate multiple agents in a decentralized manner while capturing the multi-modality of demonstrations.

    \subsection{Environments}

        \subsubsection{Two-Agent Swap}
\begin{figure}[t]
            \centering
            \begin{subfigure}[b]{0.48\linewidth}
                \centering
                \includegraphics[width=\linewidth]{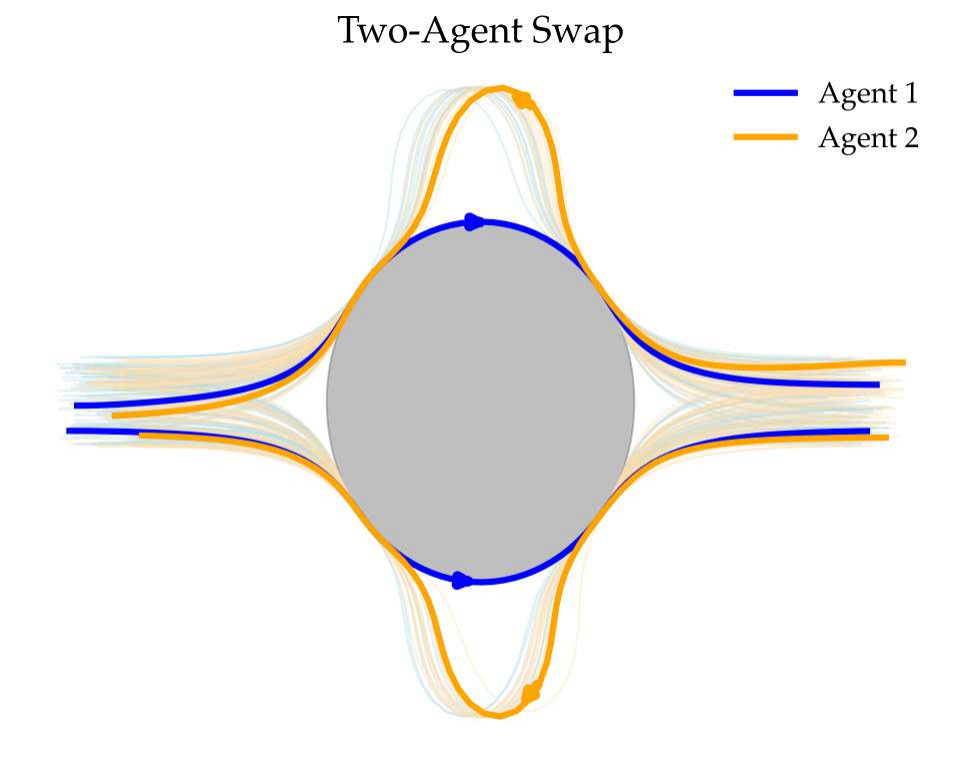}
                \caption{Two-Agent Swap}
                \label{fig:swap-env}
            \end{subfigure}
            \hfill
            \begin{subfigure}[b]{0.48\linewidth}
                \centering
                \includegraphics[width=\linewidth]{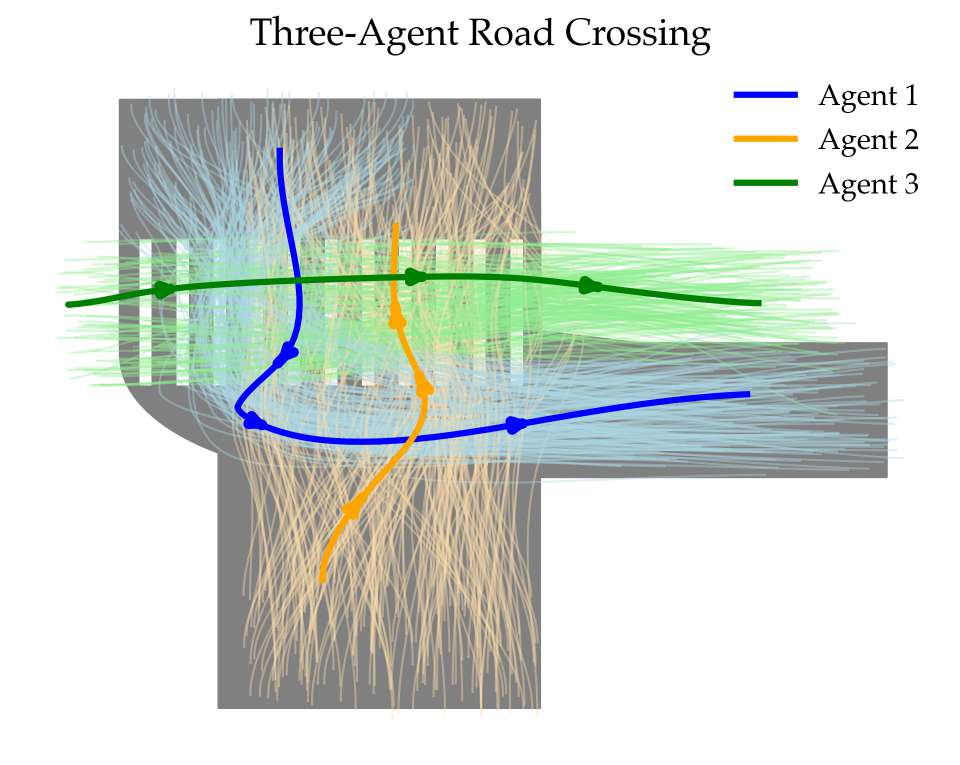}
                \caption{Three-Agent Road Crossing}
                \label{fig:road-env}
            \end{subfigure}
            \caption{
            Visualizations of the Two-Agent Swap and Three-Agent Road Crossing environments, along with examples of the multi-modal expert demonstrations used to train the various models. The Swap environment included 6 different solution modes, and the Road Crossing environment did not have explicit modes but rather more subtle collision avoidance behaviors.}
            \label{fig:sim-env}
            \vspace{-0.7cm}
        \end{figure}

        We consider an environment, which we call ``Swap,'' that involves two agents who are trying to swap positions with one another while avoiding a central obstacle on a unitless X-Y coordinate plane. This environment involves six possible modes of achieving the swapping of positions, as shown in~\cite{bhatt2025multinash}: two trivial interaction modes where both agents pass on opposite sides of the obstacle, and four nontrivial modes where both agents go on the same side of the obstacle and one yields to the other.
        A visualization of the environment and expert demonstration trajectories is shown in Fig.~\ref{fig:swap-env}. Here, coordination is crucial to ensure that agents avoid collisions with one another, and the difficulty is elevated because of how symmetric the setup is, thus making it more difficult for agents to see a ``best'' mode of operation. 
        The training dataset was evenly distributed across the six solution modes to ensure balanced coverage of the multi-modal behavior.

        \subsubsection{Three-Agent Road Crossing}
        We consider another environment called ``Road Crossing'' that involves three agents that are trying to avoid one another while on their way to their respective goal locations on a unitless X-Y coordinate plane. This environment demonstrates a need for agents to not only select a different mode of execution, as was the case in the ``Swap'' environment, but also to wait for other agents to pass before continuing to avoid collision. 
        The state and action dimensions for each agent are both two (an x-y coordinate pair), and the agents are assumed to be holonomic point robots with single-integrator dynamics. 
        We include this to test temporal coordination.
        A visualization of the environment and expert demonstration trajectories is shown in Fig.~\ref{fig:road-env}.
        Unlike the Swap environment, this setting does not have explicitly defined discrete modes; demonstrations were collected from varied collision-avoidance interactions without enforcing a predefined distribution over behaviors.

        \subsubsection{Two Arm Lift} \label{subsubsec:TwoArmLift}
        Our environment of two manipulators, based on Robosuite \cite{robosuite2020}, is shown in Fig. \ref{fig:two-arm-sim}. The two Kinova3 arms collaborate to lift the pot and transfer it to the other side while avoiding the obstacle (red box). We collect expert demonstrations in two modes: passing the obstacle on the left and the right. This is a difficult task to coordinate because the two arms need to maintain their grip on the pot without dropping or pulling on it while also deciding which way to go around the obstacle. Additionally, because the two arms are not explicitly communicating and are planning independently, making this decision while adapting to the actions of the other can be very challenging.
        For this task, expert demonstrations were evenly collected across the two modes (passing the obstacle on the left and right) to ensure balanced training data.
    \vspace{-0.1cm}
    \subsection{Baselines}
    \vspace{-0.1cm}
    We compare our performance against multiple established imitation learning baselines.
        \subsubsection{\maulik{\maulik{MLP BC}}}
        We frame this baseline as \maulik{a joint training, decentralized execution} BC. We train a separate BC policy for each agent from expert demonstrations using a shared loss function, but instead of a diffusion model, each policy is parameterized by a simple feedforward neural network. The learned policies are then executed in a decentralized manner.
        
        \subsubsection{MA-GAIL}
        MA-GAIL~\cite{song2018multi} is a GAN-based MAIL method that can be adapted to the \maulik{joint training, decentralized execution} paradigm.
        To do so, we train a single discriminator and individual generators for each agent with a shared loss.

        \subsubsection{Vanilla Diffusion}
        Typically in MAIL literature, it has been shown that matching occupancy measures of each agent individually is sufficient (see Proposition 2, ~\cite{song2018multi}).
        Therefore, we compare our approach to a simplified Diffusion variation, which we call Vanilla \maulik{Diffusion} that simply imitates each agent in the environment without accounting for other agents. This means that each agent's observation only encompasses their own state and the environment around them, but all agents' models share a loss according to \eqref{eq:total_loss} during training time. 
        Note, we do not directly compare against LatentToM and MADiff because they offer a different level of decentralization that depends on an attention mechanism that predicts states and a consensus embedding, while MIMIC-D operates on strictly local observations that achieve implicit coordination.
    \subsection{Comparison Metrics}
    We choose various comparison metrics, such as the number of collisions among agents and the success rates, to compare our method against the baselines. Furthermore, since the goal of imitation learning is to accurately capture expert behavior, in our navigation scenarios, we compare the distribution of generated trajectories with that of expert trajectories. For this comparison, we employ the Wasserstein distance~\cite{villani2009wasserstein}, also known as the Earth Mover’s Distance (EMD), a widely used metric for measuring distances between probability distributions. Computing the EMD requires defining a distance between individual elements of the distributions. Since our elements are trajectories, we use the Fréchet distance~\cite{eiter1994computing} as the underlying metric to quantify the distance between two trajectories.
        
    \subsection{Results}

        \begin{figure}[t]
            \centering
            \begin{subfigure}[b]{0.48\linewidth}
                \centering
                \includegraphics[width=\linewidth]{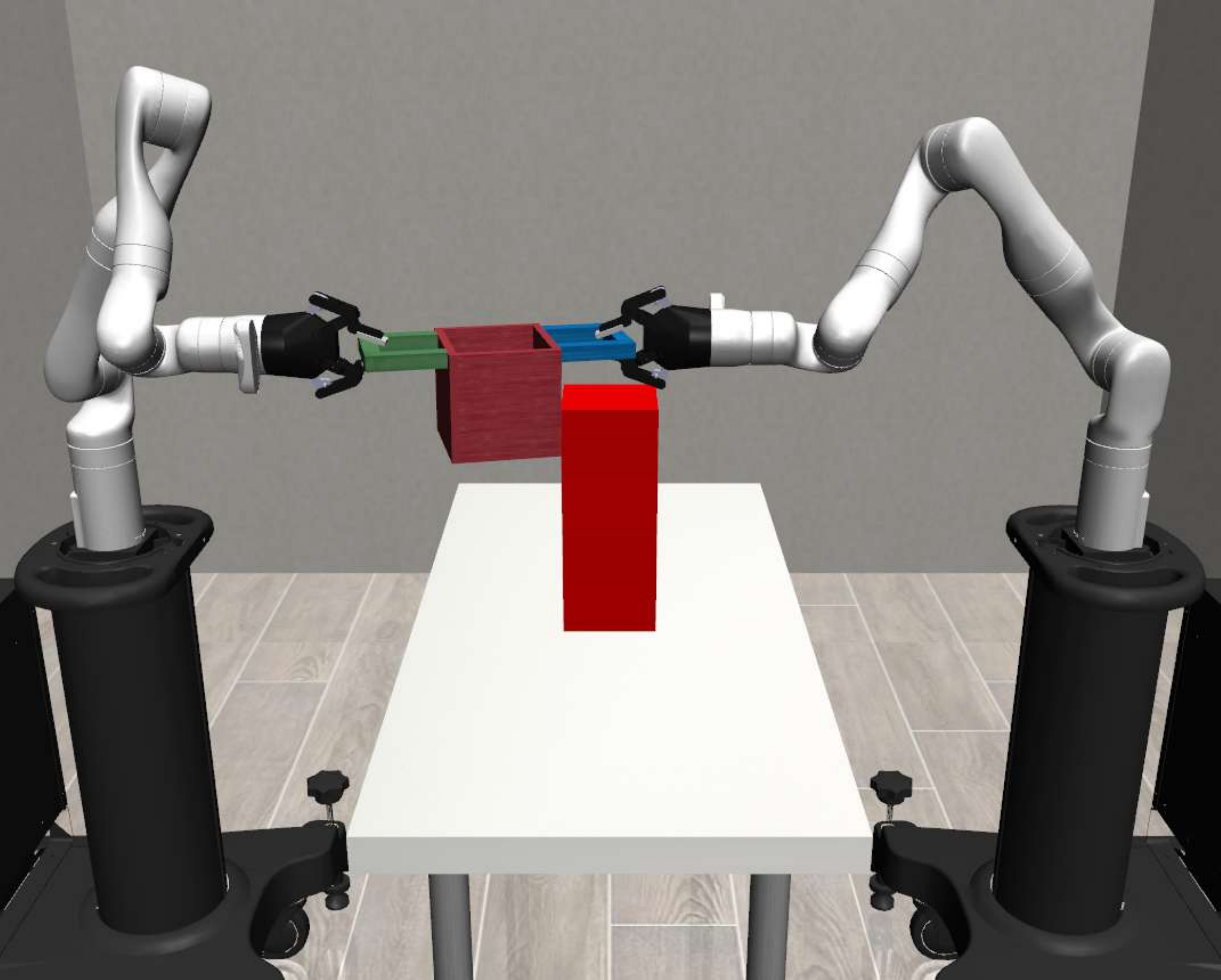}
                \caption{Two-Arm Lift Simulation}
                \label{fig:two-arm-sim}
            \end{subfigure}
            \hfill
            \begin{subfigure}[b]{0.48\linewidth}
                \centering
                \includegraphics[width=\linewidth]{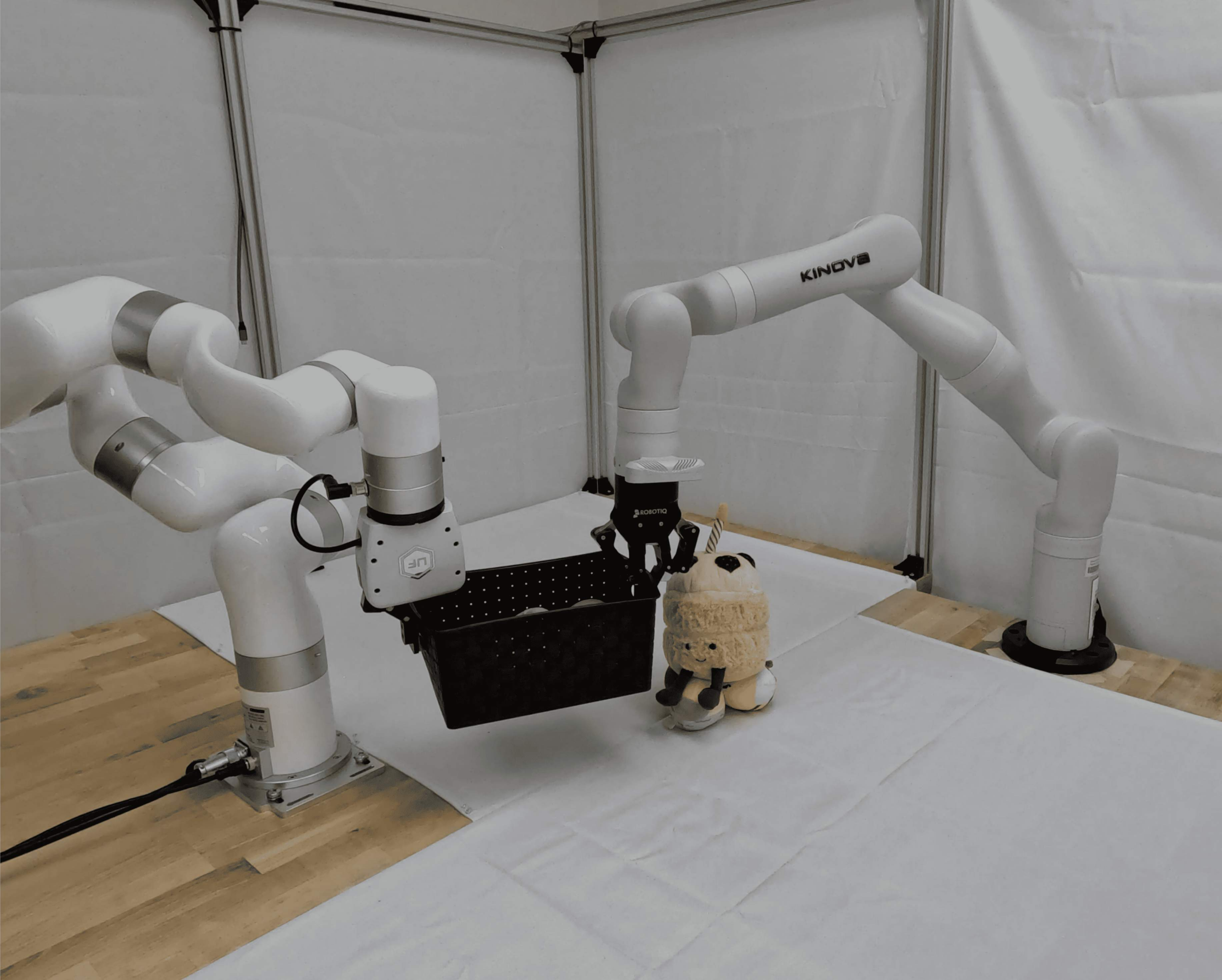}
                \caption{Two-Arm Lift Hardware}
                \label{fig:two-arm-hardware}
            \end{subfigure}
            \caption{\textbf{Two-arm lift environment visualizations.} We provide an example of what the two-arm lift task looks like for the Robosuite simulation version and the hardware demonstration. The simulation task uses two Kinova3 arms, while the hardware task uses one Kinova3 and one XArm7 arm.}
            \label{fig:bimanual-env}
            \vspace{-0.6cm}
        \end{figure}

    

        \subsubsection{Two-Agent Swap}
        In table~\ref{tab:swap_fullcollisions}, we present the number of agent-agent collisions (labeled ``Agent''), the number of agent-obstacle collisions (labeled ``Obstacle''), and the total number of collisions (either agent-agent or agent-obstacle) for 100 sampled trajectories with randomized initial positions. 
        Here, we set the observations $o^i$ of each agent to be their own position and the position of the other agent.
        In the expert demonstrations, agents maintained a separation of 3 units and an obstacle radius of 4 units. For evaluation, we use an inter-agent distance of 2.7 units and an obstacle radius of 3.9 units, respectively, to qualify a collision. This avoids counting borderline collisions caused by discretization or noise while also increasing the ability to distinguish between methods. MIMIC-D encounters the least number of total collisions by a significant margin. 
        It is worth noting that \maulik{MLP BC} encountered so few agent-agent collisions because the trajectories demonstrated mode averaging and cut through the central obstacle in nearly every instance. Vanilla Diffusion and our method both had very few agent-obstacle collisions due to diffusion's ability to effectively capture the multi-modality, but Vanilla Diffusion lacks coordination among agents, which makes it challenging for it to avoid agent-agent collisions.

        As shown by the EMD in table~\ref{tab:swap_ot} computed from 100 sampled trajectories, MIMIC-D and Vanilla Diffusion achieve the smallest values, best replicating the expert demonstration distribution. This is expected because we know that diffusion can effectively capture multi-modality without suffering from mode collapse or averaging. While both methods achieve similar EMD scores due to the shared diffusion foundation, our advantage lies in enabling inter-agent coordination, a failure point of the Vanilla Diffusion method.
        
        It is worth noting here that \maulik{MLP BC} and MAGAIL are deterministic, meaning that they would always produce a single path from a fixed initial state, unlike our method's stochastic nature, which allows it to recover the full distribution of expert demonstrations by sampling from random noise.

        \begin{table}
            \centering
            \small
            \caption{Two-Agent Swap (\# collisions / 100)
            }
            \label{tab:swap_fullcollisions}
            \begin{tabular}{lccc}
                \toprule
                Method & Agent & Obstacle & Total \\
                \midrule
                \maulik{MLP BC}             & 18  & 98  & 98 \\
                MAGAIL  & 97  & 99  & 99 \\
                Vanilla \maulik{Diffusion}  & 52  & \textbf{1}  & 52 \\
                \textbf{MIMIC-D}           & \textbf{12} & 3   & \textbf{15} \\
                \bottomrule
            \end{tabular}
            \vspace{-0.6cm}
        \end{table}

        \begin{table}[h!]
            \centering
            \small
            \caption{EMD between Expert Demonstrations and Sampled Trajectories for Two-Agent Swap}
            \label{tab:swap_ot}
            \begin{tabular}{lcc}
                \toprule
                Method & Agent-1 & Agent-2 \\
                \midrule
                \maulik{MLP BC} & 1.93 & 1.72 \\
                MAGAIL & 4.67 & 7.99 \\
                Vanilla \maulik{Diffusion} & \textbf{1.20} & 1.74 \\
                \textbf{MIMIC-D} & 1.50 & \textbf{1.24} \\
                \bottomrule
            \end{tabular}
        \end{table}

        \subsubsection{Three-Agent Road Crossing}
        
        In table~\ref{tab:road_fullcollisions}, we report agent-agent collisions for a varying collision threshold. We deem there to be a collision between two agents if the minimum distance between them at any timestep is less than the threshold. 
        Here, we set the observations $o^i$ of each agent to be their own position and the positions of the other two agents.
        Expert demonstrations were collected with a clearance of 0.75 units (the agents in the demonstrations were always at least 0.75 units apart), so we sweep the collision threshold from 0.75 to more relaxed values (0.675, 0.5625, 0.375). This range serves to distinguish true collisions from near misses near the demo threshold and also reveals how well methods learn this collision avoidance behavior. While all approaches incur many threshold hits near the conservative margin of 0.75, our approach exhibits the steepest decline as the collision threshold decreases. This indicates that our method was able to learn implicit coordination.

        Table~\ref{tab:road_ot} reports the EMD between the expert demonstrations and the 100 sampled trajectories for the various methods. From here, it is clear that MIMIC-D is able to achieve the smallest value, meaning that we best replicate the expert demonstration distribution. Once again, we note that \maulik{MLP BC} and MAGAIL are deterministic, and for a static initial position, ended up generating the same trajectory over 100 samples, while Vanilla Diffusion and our method recovered the expert trajectory distribution.

        \begin{remark}
        It may be observed that there is a potential for local minima, where robots stagnate when trying to decide on a mode similar to the Freezing Robot Problem \cite{trautman2010unfreezing}. We see in experiments with our method that robots avoid this, which we believe is because diffusion is inherently stochastic. Due to the sampling process starting from noise, trajectories have inherent randomness that assists in avoiding stalling.
        \end{remark}
        
        
        
        \begin{table}[t!]
            \centering
            \small
            \caption{Three-Agent Road Crossing (\# collisions / 100)}
            \label{tab:road_fullcollisions}
            \begin{tabular}{lcccc}
                \toprule
                Method & \multicolumn{4}{c}{Collision Threshold} \\
                \cmidrule(lr){2-5}
                       & 0.74 & 0.675 & 0.5625 & 0.375 \\
                \midrule
                \maulik{MLP BC}                    & 99  & 94  & 81  & 55 \\
                MAGAIL    & \textbf{95}  & 92  & 68  & 32 \\
                Vanilla \maulik{Diffusion}  & 96  & 93  & 85  & 61 \\
                \textbf{MIMIC-D}          & \textbf{95}  & \textbf{14}  & \textbf{1}  & \textbf{0}  \\
                \bottomrule
            \end{tabular}
            \vspace{-0.3cm}
        \end{table}

        \begin{table}[]
        \centering
        \small
        \caption{EMD between Expert Demonstrations and Sampled Trajectories for Three-Agent Road Crossing}
        \label{tab:road_ot}
        \begin{tabular}{lccc}
        \toprule
        Method & Agent 1 & Agent 2 & Agent 3 \\
        \midrule
        \maulik{MLP BC}                     & 0.4497 & 1.1541 & 0.3615 \\
        MAGAIL     & 1.2749 & 1.5166 & 0.5402 \\
        Vanilla \maulik{Diffusion}               & 0.4241 & 0.5046 & 0.3301 \\
        \textbf{MIMIC-D}                   & \textbf{0.3259} & \textbf{0.4394} & \textbf{0.2845} \\
        \bottomrule
        \end{tabular}
        \vspace{-0.5cm}
        \end{table}


        \subsubsection{Two Arm Lift} \label{subsubsec:two_arm_lift}
        We used 50 expert demonstrations per mode to train our model. The action trajectory length is 25, and we only execute the first 10 steps of the planned trajectory. We use the end-effector pose controller provided in Robosuite to track the planned trajectory.
        Here, we set the observations $o^i_t$ of each agent $i$ at time $t$ to be the end-effector pose of the ego agent, the end-effector pose of the other agent, and the pot's initial handle positions. Note that the poses are expressed in the respective local base frames, and it is up to the policy to interpret the other agent's pose. 

        After the training, we randomly spawn the pot into the environment and test our method and baselines on the lift task. For the lift task, two important sub-tasks require robot coordination to achieve the task successfully:
        \begin{itemize}
            \item The robots need to reach the pot handle and simultaneously lift the pot. If one robot tries to move fast without waiting for the other, it may fail to lift the pot, which would result in the entire task failing.
            \item After they successfully lift the pot, the robots need to move in coordination and choose the same mode in order to successfully transport the pot.
        \end{itemize}

        Therefore, we report the success rates of MIMIC-D and the baselines on these two subtasks for 20 different runs in the table~\ref{tab:lift_sim}. As shown in table~\ref{tab:lift_sim}, our method almost always picks up the basket even when placed randomly, and ours is the only method that is able to successfully transport the basket. 
        MAGAIL performs the worst due to the inherent instability of GAN training, which prevents it from learning reliable policies for this complex, high-dimensional task, especially when coordination is required. \maulik{MLP BC} achieves partial success in lifting the pot, but the requirement of coordinated behavior between agents makes it difficult for \maulik{MLP BC} to complete the full task. Vanilla Diffusion benefits from the strengths of diffusion models and learns higher-quality policies than \maulik{MLP BC}, but still fails to capture inter-agent coordination in the presence of multi-modality. In contrast, MIMIC-D effectively leverages diffusion and enables our method to consistently lift and transport the pot with the highest success rate.


        \begin{table}[h!]
        \centering
        \small
        \caption{Two-Arm Lift simulation success rates (20 runs)}
        \label{tab:lift_sim}
        \begin{tabular}{lccc}
        \toprule
        Method & Number of &  Number of \\
        & successful lifts & overall success \\
        \midrule
        \maulik{MLP BC} & 2 & 0 \\
        MAGAIL & 0 & 0 \\
        Vanilla \maulik{Diffusion} & 5 & 0 \\
        \textbf{MIMIC-D}  & \textbf{18} & \textbf{15} \\
        \bottomrule
        \end{tabular}
        \vspace{-0.3cm}
        \end{table}

        
    \section{Hardware Experiments} \label{sec:hardware_experiments}
    To demonstrate MIMIC-D in high-dimensional real hardware, we also perform the Two-Arm Lift task in the real world. The experiment setup is shown in Fig.~\ref{fig:two-arm-hardware}. We use one xArm7 arm and one Kinova3 arm to lift and transfer the basket while avoiding an obstacle. This setup demonstrates the ability of our approach to handle heterogeneity among the agents.
    We simplify the setting by fixing the pose of the pot and removing it from the observation vector in Section~\ref{subsubsec:TwoArmLift} since we already demonstrated MIMIC-D's ability to complete this task. Here, we collect two modes of expert demonstrations (left and right) with only eight demonstrations per mode. 
    The xArm7 tracks the planned pose trajectory with an end-effector pose controller, while the Kinova3 tracks the trajectory with an end-effector velocity controller according to \cite{seo2023geometric}.

    During execution, the two arms need to agree on which mode to execute. Despite initial attempts to move in opposite directions, after replanning, they reach a consensus. In Fig~\ref{fig:fig1}, a full demonstration of both modes is shown with the decision point occurring at $t_2$. This coordination is very difficult and demonstrates that the agents can understand both modes and adapt online only through implicit coordination. In our experiments, our method achieved \textbf{95\% success rate} (19 successes out of 20 trials), demonstrating our method's reliability even when trained with only 16 total demonstrations. Although the arms show some oscillatory behavior in choosing a mode, successful attempts ultimately converge and place the basket on the opposite side of the obstacle. 

    We observe a higher success rate in hardware compared to simulation, primarily because in simulation, the pot is modeled as a rigid body, meaning even small oscillations could cause drops, while the hardware setup uses a slightly deformable basket that is more tolerant to oscillations. It is important to note that our method achieves this high success rate by reliably completing the task, not by exploiting the deformability. As we mentioned in Section~\ref{subsubsec:two_arm_lift}, the failure points in simulation for baseline methods are much more severe and occur before oscillations affect the task success. Therefore, these methods would only perform as well as they did in the simulation.

%% file: conclusion.tex
\vspace{-0.2cm}
\section{Conclusion} 
\vspace{-0.1cm}
\label{sec:conclusion}
We introduced a novel multi-agent imitation learning approach called MIMIC-D for decentralized multi-agent coordination in multi-modal tasks. By formulating the problem as a dec-POMDP and modeling agent policies as a conditional Diffusion Transformer model, we were able to recover diverse multi-model behaviors and avoid failure points like mode collapse. Our paradigm overcomes the need for agents to use a centralized planner to coordinate among agents using only locally available information.
We demonstrate the effectiveness of this approach in multiple simulated domains and on a complicated bimanual hardware setup. In all these setups, we show significant improvements over baselines in recovering expert trajectory distributions while reducing collisions and task failures.
Despite these improvements, our approach exhibits limitations in highly symmetric environments, where decentralized agents may independently select incompatible modes, leading to collisions or stagnation during mutual avoidance.
In the future, we aim to remove the assumption of perfect state observation by conditioning on camera inputs. Also, we will look into human-robot interactions by operating one of the robots manually and allowing others to plan autonomously. We will also explore ways to improve out-of-distribution performance, a common shortcoming of IL methods.